\documentclass{article}

\PassOptionsToPackage{numbers}{natbib}

\usepackage[preprint]{neurips_2025}

\usepackage[utf8]{inputenc} 
\usepackage[T1]{fontenc}    
\usepackage{hyperref}       
\usepackage{url}            
\usepackage{booktabs}       
\usepackage{amsfonts}       
\usepackage{nicefrac}       
\usepackage{microtype}      
\usepackage{xcolor}         
\usepackage{multirow}       
\usepackage{graphicx}       
\usepackage{subcaption}
\usepackage{placeins}

\usepackage{changes}

\pdfoutput=1

\title{Automatic Dataset Generation for Knowledge Intensive Question Answering Tasks}

\author{%
Sizhe Yuen$^{1}$ \quad Ting Su$^{1}$ \quad Ziyang Wang$^1$ \quad Yali Du$^{1,2}$ \quad Adam J. Sobey$^{1,3}$ \\
$^1$The Alan Turing Institute \quad $^2$King's College London \quad $^3$University of Southampton\\
\texttt{\{syuen,tsu,zwang,asobey\}@turing.ac.uk}\\
\texttt{\{yali.du\}@kcl.ac.uk}
}

\begin{document}

\maketitle

\begin{abstract}
A question-answering (QA) system is to search suitable answers within a knowledge base. Current QA systems struggle with queries requiring complex reasoning or real-time knowledge integration. They are often supplemented with retrieval techniques on a data source such as Retrieval-Augmented Generation (RAG). However, RAG continues to face challenges in handling complex reasoning and logical connections between multiple sources of information. A novel approach for enhancing Large Language Models (LLMs) in knowledge-intensive QA tasks is presented through the automated generation of context-based QA pairs. This methodology leverages LLMs to create fine-tuning data, reducing reliance on human labelling and improving model comprehension and reasoning capabilities. The proposed system includes an automated QA generator and a model fine-tuner, evaluated using perplexity, ROUGE, BLEU, and BERTScore. Comprehensive experiments demonstrate improvements in logical coherence and factual accuracy, with implications for developing adaptable Artificial Intelligence (AI) systems. Mistral-7b-v0.3 outperforms Llama-3-8b with BERT F1, BLEU, and ROUGE scores 0.858, 0.172, and 0.260 of for the LLM generated QA pairs compared to scores of 0.836, 0.083, and 0.139 for the human annotated QA pairs.
\end{abstract}

\section{Introduction}
QA systems are designed to provide relevant answers to user questions. They are increasingly being developed for customer support chatbots, summaries of technical documents or generating reports. The advent of deep learning has significantly advanced QA systems through sophisticated neural network architectures such as Transformers \citep{vaswani2017attention,devlin2018bert}, improving their ability to understand and generate responses. However, these models often struggle with queries requiring extensive factual knowledge or real-time data, due to a crucial dependency on external knowledge sources. To address this issue, \citet{lewis2020retrieval} introduced RAG, which merges the depth of traditional Information Retrieval (IR) techniques with the generative capabilities of advanced language models. RAG demonstrated effectiveness in integrating real-time retrieval into the generation process, enabling models to dynamically access and utilise relevant external documents. RAG models combine the strengths of generative pre-trained transformers, such as GPT-3 \cite{brown2020language}, with robust document retrieval systems, marking a significant advancement in knowledge-intensive QA tasks.

Despite these advancements, language models for QA face challenges in providing factual answers and logically sound reasoning to questions, particularly for queries which require broad knowledge and complex reasoning chains to deliver precise, context-aware responses. Traditionally, open-domain QA systems relied on keyword matching and shallow parsing techniques, limiting their reasoning capabilities \cite{chen-etal-2017-reading}. While RAG demonstrated effectiveness in integrating real-time retrieval, it still faces challenges in complex reasoning tasks that require synthesising information from multiple sources and performing multi-step inferences \cite{tang2024multihoprag}. The core limitations of RAG include retrieval inaccuracies, document consolidation errors, generation hallucinations, and high retrieval latency issues \cite{gao2024rag-survey,barnett2024ragfailurepoints}. Current research focus on developing adaptive weighting schemes for balancing retrieved and intrinsic information \cite{wang2024astuteragovercomingimperfect}, and exploring meta-learning approaches to improve the model's reasoning capabilities when reconciling conflicting data points \cite{xiang20252metareasoningllmslearning}. Furthermore, retrieving and reasoning over contextual information from large-scale document collections presents significant challenges in maintaining logical coherence, as interconnections among documents require not just retrieval but also logical inference to establish relationships between concepts \cite{gao2024rag-survey}. Supervised fine-tuning is a method to alleviate these issues and improve model performance, teaching the models to reason and learn information in a specific domain which comes at a higher computational cost of training and requires labelled datasets \cite{balaguer2024ragvsfinetuningpipelines}. Unsupervised fine-tuning does not require labelled dataset but has been shown to be outperformed by RAG solutions \cite{ovadia2024finetuningretrievalcomparingknowledge} by between 3.5\% and 6\%.
The cost of training and creation of labelled datasets highlights a need for automatic dataset creation methods to reduce the cost of supervised fine-tuning for LLM-based QA tasks.

In this paper, we propose a novel approach to reduce the cost of fine-tuning QA systems by introducing a self-improving cycle that creates context-based QA pairs from external documents, which are then used to fine-tune an LLM.
This approach aims to reduce the need for human labelling in 
knowledge-intensive QA tasks, accelerating the pace at which LLMs can adapt to new domains and information by enhancing both knowledge integration and reasoning capabilities in LLMs. We introduce a two-stage prompt engineering technique with two primary components: (1) an automated QA generator, which creates contextually relevant and diverse QA pairs from given documents and (2) a Model Fine-Tuner that leverages the generated QA pairs to iteratively update the target LLM, gradually expanding its knowledge base and improving its comprehension capabilities. This framework enhances the model's comprehension, but also ensures that generated responses are grounded in the provided context to maintain fidelity to the source material. 

In summary, our contribution is a novel method to enhance LLMs for knowledge intensive QA tasks with an automated approach to generate synthetic QA pairs from contextual documents. We provide empirical evidence for the effectiveness of our method through comprehensive evaluation metrics, particularly focusing on the quality of logical reasoning in generated responses. Our approach demonstrates how LLMs can be leveraged to create the training data that captures both factual knowledge and reasoning patterns, potentially opening new avenues for developing more sophisticated reasoning capabilities in AI systems.

\section{Related Work}

We first explore the related work in RAG, the strengths and limitations of its approaches. Then we discuss supervised fine-tuning, an alternate method to RAG for improving LLM performance and expertise in specific domains. This highlights a need for synthetic data generation, to reduce the cost of creating labelled datasets for fine-tuning.

\textbf{RAG} integrates IR techniques with generative models to enhance the accuracy and relevance of generated answers. Specifically, it generally involves two interlocking modules: the retriever, which aims to gather relevant documents based on the question, and the generator, which interprets both the question and the relevant documents to form comprehensible answers. The retrieval component, often based on Dense Passage Retrieval (DPR) \cite{karpukhin2020densepassageretrievalopendomain}, fetches relevant documents from a large corpus like Wikipedia. The generative component, typically a sequence-to-sequence model like BART~\cite{lewis2020bart}, generates answers conditioned on the retrieved documents. \citet{ovadia2024finetuningretrievalcomparingknowledge} introduced a comprehensive RAG framework that utilizes a dense vector index of Wikipedia for document retrieval and a BART model~\cite{lewis2020bart} for answer generation. This model was demonstrated to outperform extractive models by generating more comprehensive and contextually accurate answers, even when the correct answer is not in any retrieved document.

Subsequently, researchers developed many RAG-based models~\cite{chalkidis2020legal,jeong2024adaptive,ye2024boosting,glass2022re2g,asai2024selfrag,lin2024radit,shi2024replug} that leverage and improve upon the original RAG model. 
For instance, in the legal field,~\citet{chalkidis2020legal} proposed the CBR-RAG model, which integrates case-based reasoning to enhance retrieval, making it particularly suitable for legal QA tasks. 
In conversational QA, a fine-grained retrieval augmentation approach~\cite{ye2024boosting} has been proposed to refine question understanding and improve the relevance of retrieved information by question reformulation and keyword extraction to better align the retrieved documents with the user's query. 
In addition,~\citet{glass2022re2g} aims to improve the ranking of retrieved information by introducing a BERT-based reranker trained by additional high-quality labelled data.

However, the effectiveness of a RAG system heavily depends on the accuracy of its retrieval mechanism, which itself is dependent on the accurate chunking and tagging of relevant documents. Complex or ambiguous queries may lead to retrieval of irrelevant
or misaligned document chunks, leading to incomplete or incorrect answers \cite{gao2024rag-survey}. These errors are particularly pronounced in naive
implementations where the document context is simply prepending to an LLM prompt \cite{izacard2021leveragingpassageretrievalqa}. More advanced RAG techniques of pre-retrieval \cite{ma2023queryrewritingrag,gao2022precisezeroshotdenseretrieval} and post-retrieval strategies further improves on naive RAG implementations at the cost of addition complexity and resources, which may exacerbate the latency issues of RAG approaches. 

\textbf{Supervised Fine-tuning} has been another approach shown to improve the performance of LLMs in QA scenarios, either in isolation or in conjunction with RAG architectures. For example, the DPR used in RAG can be fine-tuned within a RAG system to improve alignment between retrieved documents and LLM responses \cite{siriwardhana2021finetuneentireragarchitecture}.
In addition to training better retrievers to enhance the RAG model in QA tasks, researchers also aim to fine-tune the language model to produce more relevant and coherent answers for questions based on the retrieved context. \citet{asai2024selfrag} introduces the Self-RAG framework, training a language model to retrieve, generate, and critique its own outputs through self-reflection. Another approach is to separately fine-tune the language model and the retrieval models. The RA-DIT framework \cite{lin2024radit} introduces a fine-tuning methodology to separate the two fine-tuning steps, first fine-tuning the language models to better utilise retrieved information, then fine-tuning the retriever to return results guided by the language model preferences. When fine-tuning just the retrieval model, researchers have also treated the language model as a black box and simply prepend the retrieved documents to the language model inputs \cite{shi2024replug}. 
There are two major disadvantages to the fine-tuning approach: the cost of training, and the requirement of a labelled dataset. The former can be alleviated by Parameter-Efficient Fine-Tuning methods such as Low Rank Adaptation \cite{hu2021loralowrankadaptationlarge} and quantisation \cite{dettmers2023qlora}. Our work aims to solve the the latter through the creation and use of synthetic datasets in the fine-tuning process.

\textbf{Synthetic Dataset Generation}. 
The use of LLMs for synthetic data generation has been gaining increasing attention as a solution to reduce the cost of human-labelling datasets. This approach has been shown to be successful across a number of domains, such as text classification \cite{li2023syntheticdatageneration},
human-computer interaction (HCI) \cite{hamalainen2023synthetic-hci}, 
and code generation \cite{nadas2025syntheticdatageneration}. Some of these examples, such as the HCI case study by \citet{hamalainen2023synthetic-hci}, where the generated dataset is questionnaire responses on experiencing video games as art, do not use the generated dataset for further LLM training.

In the QA domain, one synthetic data generation approach is dialogue inpainting, where text from a given document is transformed into a two-person dialogue between a writer and reader \cite{dai2022dialoginpainting}. These existing approaches demonstrate the effectiveness of synthetic data generation, but have yet to be applied to QA generation in a technical setting. We take a similar approach, but instead of transforming a given document into dialogue, we prompt the model to generate relevant questions and answers with the given document in technical industrial domains, where there is an abundance of documentation, but the cost to create a human-labelled dataset is high and requires some level of expertise in the domain.

\section{Automatic Dataset Generation Methodology}
\begin{figure}[htb]
  \centering
  \includegraphics[width=1\linewidth]{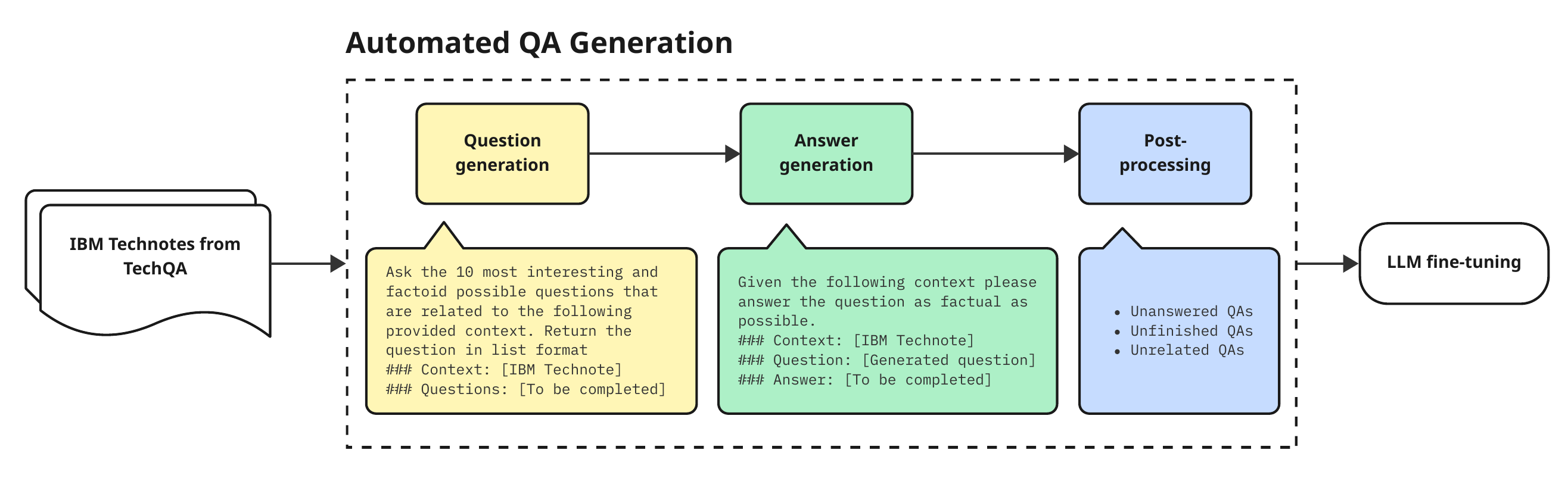}
  \caption{Example generation procedure of QA pairs with the TechQA dataset.}
  \label{fig:generation_procedure}
\end{figure}

Our approach leverages a self-improving cycle to produce a synthetic dataset that minimizes human intervention while maximizing the potential for continuous learning and adaptation. The code for our generation and fine-tuning processes is provided as a supplementary file. Figure~\ref{fig:generation_procedure} illustrates the system architecture and data flow, which comprises of two main components:

\begin{enumerate}
    \item Automated QA Generator - An LLM fine-tuned to generate relevant QA pairs from the initial dataset using three steps: question generation (yellow), answer generation (green), and post-processing (blue).
    \item Model Fine-Tuner - Utilizes generated QA pairs to update the target LLM using standard fine-tuning approaches.
\end{enumerate}

\subsection{Automated QA Generator}
For automatically generating a large quantity of QA pairs, we use Mistral-7b-instruct-v0.3 hosted on Replicate\footnote{\url{https://replicate.com}} as our baseline language model.  Mistral-7b-instruct-v0.3 is a small open-source language model that performs well on recent LLM benchmarks \cite{jiang2023mistral}. The QA generation is conducted in three steps: 

\textbf{Question Generation} -- An LLM, in these experiments Mistral, generates a set of questions for each context document given in the dataset. 
A maximum of 10 factoid question per context document are requested from any given dataset, such as technical documentation. These are generated using the prompt in the yellow box in Figure~\ref{fig:generation_procedure}.

\textbf{Answers Generation} -- Next, for each question related to each context document, we use another prompt to ask the model to generate a factual answer. This creates up to 10 question answer pairs per document. These are generated using the prompt in the green box in Figure~\ref{fig:generation_procedure}. Note that the LLM is allowed to answer "There are no possible factual answers based on the given content." for any question deemed unanswerable. These questions are subsequently removed from the dataset during post-processing. 

\textbf{Post-Processing} -- Finally, we clean up the generated dataset, which now would contain 10 QA pairs per context document used during the generation stages. We use several methods to clean up the dataset, as follows:
\begin{itemize}
    \item \textbf{Unanswered QA Check.} Although we have directly told the LLM to answer unanswerable questions using the response "There are no possible factual answers based on the given content.", the LLM sometimes answers unanswerable questions in a different way. To combat this, we use a BERT-based semantic similarity score to identify unanswered questions and remove the QA pair from the dataset. 
    \item \textbf{Unfinished QA Check.} Some answers are unfinished, leading to partially answered questions. We use a Roberta-based sentence completion classification model to identify unfinished sentences. We do not remove these QA pairs from the dataset, but instead label them with ``***unfinished***" at the end of the answer. Note that we can only identify unfinished sentences but not the entire answers. Thus if an answer is unfinished, but the last sentence in the answer is finished, this answer is considered finished. 
    \item \textbf{Unrelated QA Pair Check.} Some of the generated answers are not related to the question or the provided context. We use a BERT-based semantic similarity score to identify unrelated QA pairs and automatically remove them from the dataset. 
\end{itemize}

\subsection{Fine-Tuning Process}
An LLM is then fine-tuned using the dataset, two models are selected for comparison: Llama-3-8b  \cite{grattafiori2024llama3herdmodels} and Mistral-7b-v0.3 \cite{jiang2023mistral7b}. The Llama-3-8b model is licensed under the Meta Llama 3 community license, and Mistral-7b-v0.3 is licensed under Apache 2.0. Both models are fine-tuned with 4-bit quantisation using QLoRA~\cite{dettmers2023qlora} for efficiency and reduced memory usage, targeting all linear modules, with rank $r = 8$ and $\alpha = 16$. The 4-bit quantisation and low rank $r$ allows us to fit both the Llama 3 and Mistral models for fine-tuning. An 8-bit AdamW optimiser is used with a weight decay of $0.01$ and learning rate of 5e-5 as smaller learning rates and weight decay values are recommended with QLoRA on small models \cite{dettmers2023qlora}.
Furthermore, the Unsloth \cite{unsloth} versions of the Llama 3 and Mistral v0.3 models are used to improve performance and reduce memory usage. Contextual documents are provided in fine-tuning with the following prompt:

\noindent\fbox{%
    \parbox{\textwidth}{%
        \texttt{Find the answer to the question in the given document.}
        
        \texttt{\#\#\# Question: [Question] }
        
        \texttt{\#\#\# Document: [Document]}
    }%
}

A 0-shot, 1-shot, and 5-shot prompting strategy is
used during the fine-tuning process, as few-shot prompts have been demonstrated to improve language model performance \cite{brown2020languagemodelsfewshotlearners}.

\subsection{Evaluation Metrics}
We evaluate our approach through a comprehensive framework of automated metrics. For generation quality, we employ perplexity measurements~\cite{jelinek1977perplexity}, ROUGE scores~\cite{lin2004rouge} and BLEU scores~\cite{papineni2002bleu}, alongside a semantic similarity based BERTScore~\cite{Zhang2020bertscore}. 

\textbf{Perplexity} provides insight into the model's predictive capabilities on unseen data. 
\textbf{ROUGE score} metrics evaluate the quality of generated answers against the reference answers to provide the semantic and structural similarities between generated and reference answers. 
\textbf{BLEU scores} complement ROUGE by assessing the fluency and adequacy of generated answers.
\textbf{BERT Score} leverages contextual embeddings from the BERT model \cite{devlin2018bert} to compute similarity scores between generated and reference texts to capture deeper semantic similarity beyond surface-level token matching. We report precision, recall, and F1 from the BERT Score.
Through ablation studies, we analyse the contribution of each component in our pipeline.

\section{Performance of Automatic Dataset Generation with Fine-Tuning on TechQA dataset}

The TechQA dataset~\cite{castelli2020-techqa} is selected for training and evaluation. It is a domain-specific QA dataset tailored for technical support, distributed under the Apache 2.0 license. The QA pairs were taken from the IBM technical support forums, with answers that are linked to specific IBM technical documents (IBM Technotes). The dataset therefore presents a realistic QA scenario with real internal documentation. It is chosen over more popular general knowledge QA datasets such as HotpotQA~\cite{yang2018hotpotqa}, TriviaQA~\cite{joshi2017triviaqa}, and Natural Questions~\cite{alberti2019naturalquestions} as we aim to target LLM training in scenarios where internal documentation is used. Datasets including HotpotQA and Natural Questions are based on data from Wikipedia, which has a high risk of data contamination from pre-training, leading to a lack of domain-specific performance. The use of internal technical documentation is more closely aligned to real-world engineering applications where a QA system is trained on private internal documentation unseen by general LLM pre-training.

Table~\ref{tab:generated-qa} presents the statistics of the generated QA dataset from the TechQA context documentation. Using 11,960 IBM Technotes as context documents, on average we are able to generate 4.6 QA pairs per document, creating a total of more than 50,000 QA pairs after removing unfinished pairs in the post-processing. This is compared to TechQA's 1,400 QA pairs which were generated by five professional annotators and a sixth who was a Linux system administrator, requiring two weeks of training before the annotation period began.

\begin{table}
  \centering
  \caption{Statistics on generated QA dataset based on TechQA dataset.}
  \label{tab:generated-qa}
  \begin{tabular}{c||c|c|c}
  \toprule
    & \# of  & average \# of  & total \# of   \\ 
        & context documents &  QAs/context &  generated QAs  \\ \hline
    All QA & 11,960 & 4.614 & 55,179 \\ 
    Unfinished QA & 2,389 & 1.469 & 3,509 \\
    \bottomrule
  \end{tabular}
\end{table}

The Llama-3-8b and Mistral-7b-v0.3 models are fine-tuned and tested against the TechQA dataset. The original training and development test splits are maintained from TechQA. We test each model in three fine-tuning set-ups: no training, training on the original dataset, and training on the synthetic generated dataset. All fine-tuning set-ups provide the contextual documents to the model. For each different training set used, we train and evaluate under 0-shot, 1-shot, and 5-shot prompts. We test under two experimental settings for inference. The first where the task is assumed to be small or important enough so that the human QA pairs cover the whole space of interest, and the specific contextual document which contains the answer to the questions are provided to the model in the prompt. This tests the models' ability to understand and find the answer within the given document. However, this requires an expert understanding of the technical documents and the time to find the correct document. In the second experimental setting, it's assumed that the contextual document is not provided as the user is unfamiliar with the material. This tests how well the models have learned and retained knowledge from the fine-tuning process, and generalising QA beyond human-derived pairs. Tables~\ref{tab:techqa} and~\ref{tab:techqa-no-context} in the appendix display the results in full for fine-tuning with and without context respectively, the best results are highlighted in bold. All experiments were conducted on an A100 GPU with 40GB of memory, using the (anonymized due to double-blind policy) HPC service.

\subsection{With Context}
\begin{figure}[tbh]
  \centering
     \begin{subfigure}[b]{0.49\textwidth}
         \centering
         \includegraphics[width=\textwidth]{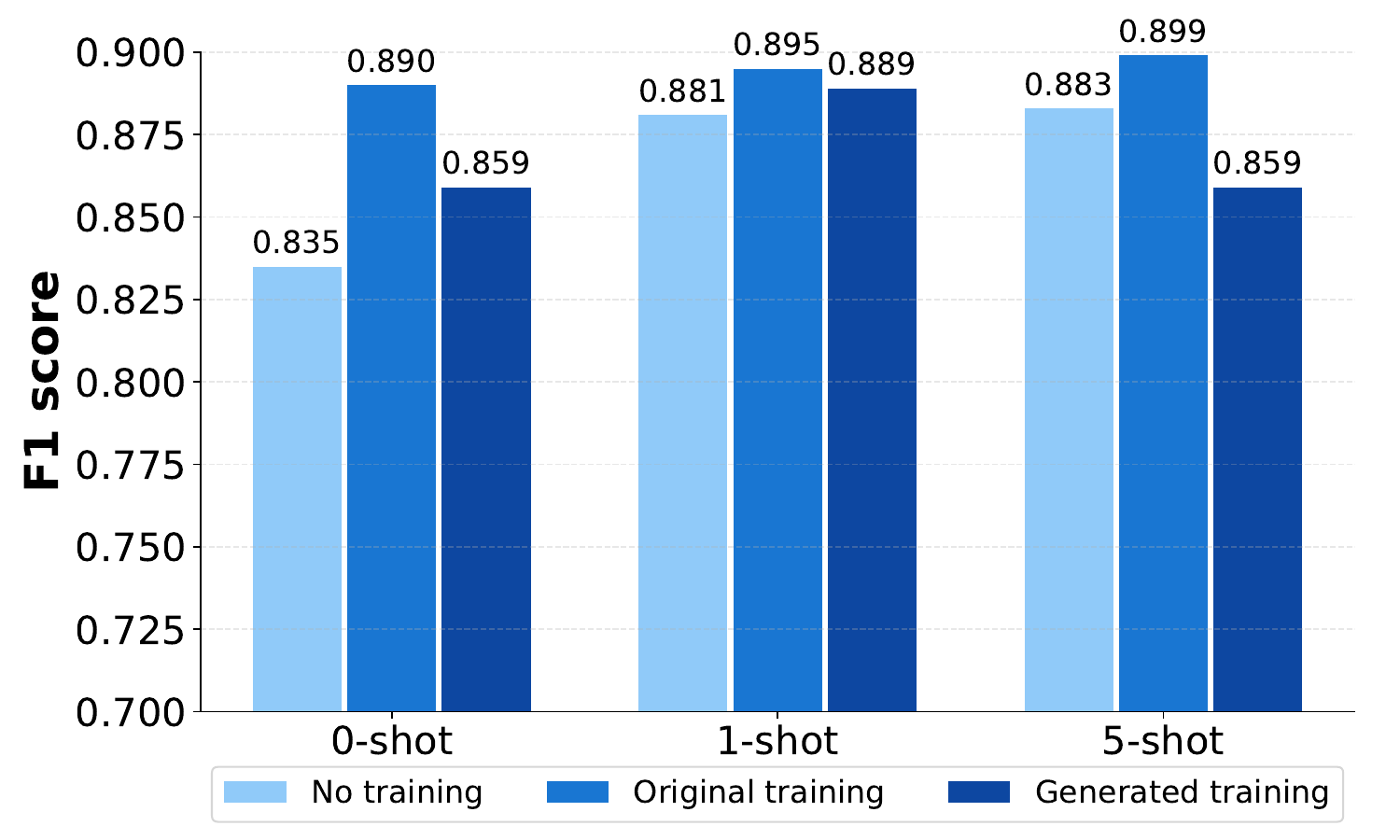}
         \caption{Llama-3-8b}
         \label{subfig:llama-with-context}
     \end{subfigure}
     \hfill
     \begin{subfigure}[b]{0.49\textwidth}
         \centering
         \includegraphics[width=\textwidth]{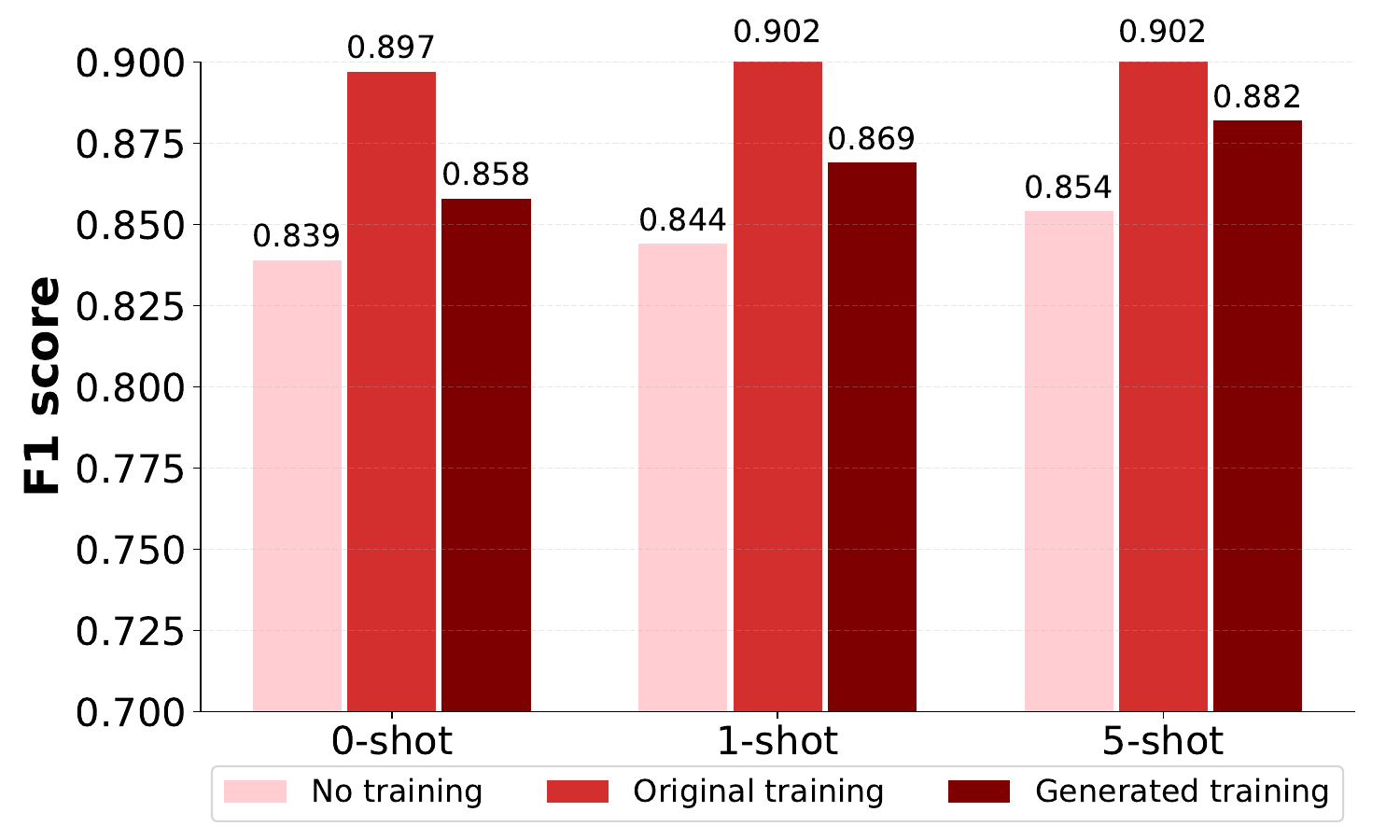}
         \caption{Mistral-7b-v0.3}
         \label{subfig:mistral-with-context}
     \end{subfigure}
  \caption{F1 score when answering with the context document provided.}
  \label{fig:f1-with-context}
\end{figure}

When the models are provided the contextual documents in the fine-tuning, the specific document that contains the answer to the questions, training on the original dataset shows the best performance. Figure~\ref{fig:f1-with-context} displays the best F1 scores of Llama-3-8b and Mistral-7b-v0.3 in 0-shot, 1-shot, and 5-shot training. The F1 scores are shown here as it comprises of both the precision and recall of the answers, though the same trend is seen in the BLEU and ROUGE scores. Training on the original dataset shows the best improvement on both models. Mistral is the most consistent across all three prompting strategies, with no training displaying the worst performance and original training providing the best performance.

Furthermore, 0-shot training for Mistral performs the worst and 5-shot training performs the best, with 1-shot and 5-shot training using the original dataset demonstrating the same F1 score. With 5-shot training, Llama-3-8b achieves an F1 score of 0.899, a BLEU score of 0.489 and a ROUGE score of 0.494 and Mistral-7b-v0.3 achieves an F1 score of 0.902, a BLEU score of 0.463, and a ROUGE score of 0.485. Llama shows slight inconsistency between prompting strategies. For example, the F1 scores for 1-shot training are similar between each training dataset, and in the 5-shot training, the generated dataset performs worse than the original dataset.
 
In general, training with the original dataset provides the best performance across all three prompting strategies, and training with the generated dataset improves performance against the baseline model without training.

The best results coming from training with the original dataset shows that generated datasets are outperformed by human-labelled datasets for finding answers given the correct document. This assumes the correct document is fetched by a retrieval system for the LLM or that the user is familiar with the material. Using the generated synthetic dataset is able to improve upon the original model without training, but is outperformed when training on the original dataset, indicating both the advantage and limitation of using the synthetic dataset.

\subsection{No Context}
When fine-tuning and inference are performed without the contextual documents, the models' knowledge and reasoning is tested. Model performance is generally lower as expected, since the models are not provided the technical documentation which contains the answers to the questions. However, we find that when no context is provided, using the generated dataset on both Llama and Mistral models exhibits the best performance in 1-shot and 5-shot prompts.

\begin{figure}[tbh]
  \centering
     \begin{subfigure}[b]{0.49\textwidth}
         \centering
         \includegraphics[width=\textwidth]{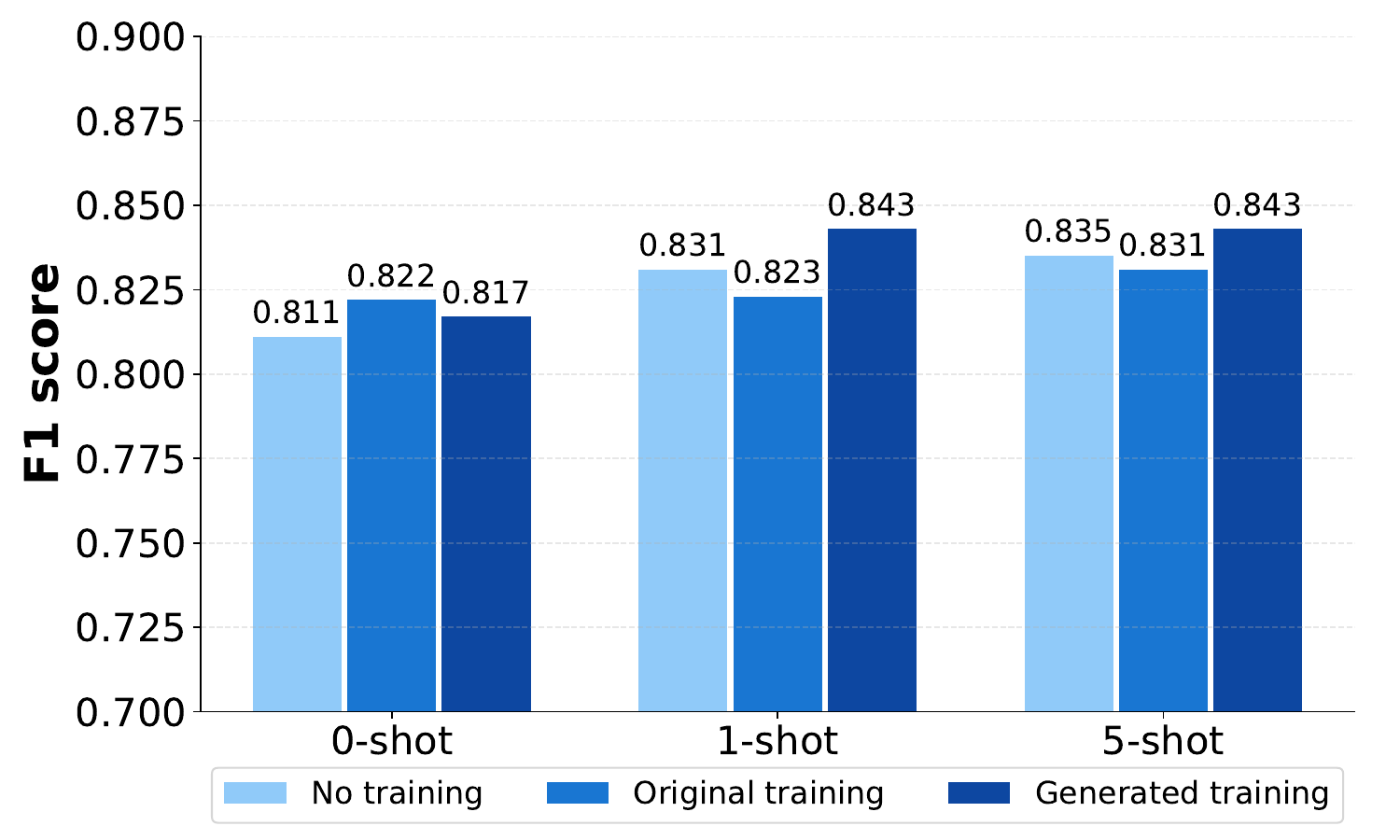}
         \caption{Llama-3-8b}
         \label{subfig:llama-no-context}
     \end{subfigure}
     \hfill
     \begin{subfigure}[b]{0.49\textwidth}
         \centering
         \includegraphics[width=\textwidth]{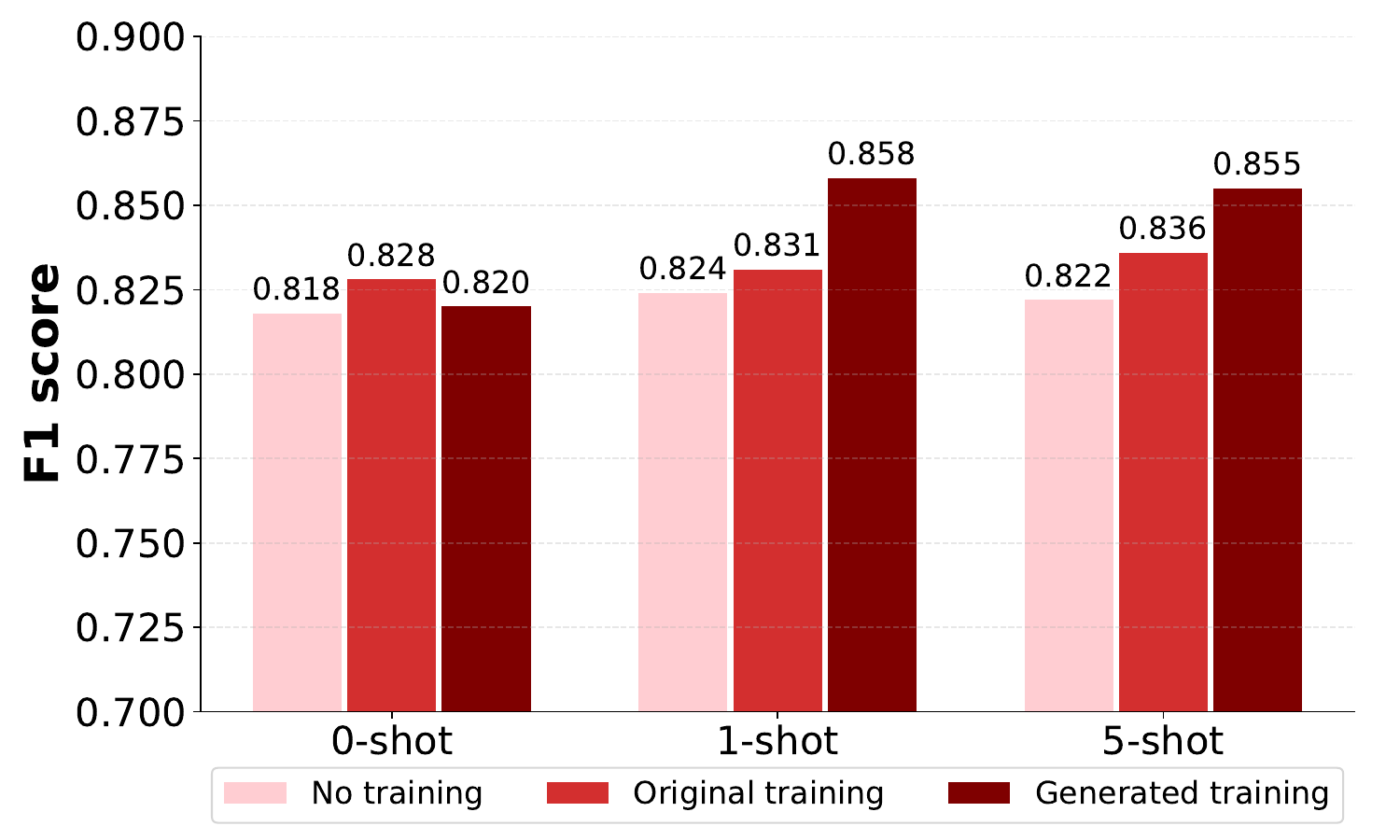}
         \caption{Mistral-7b-v0.3}
         \label{subfig:mistral-no-context}
     \end{subfigure}
  \caption{F1 score when answering with no context provided.}
  \label{fig:f1-no-context}
\end{figure}

With the exception of the 0-shot prompting strategy, Figure \ref{fig:f1-no-context} shows the use of the generated training set leads to higher F1 scores on both the Llama and Mistral models. Furthermore, the Llama model demonstrates a lower F1 score on 1-shot and 5-shot prompts compared to the base model before fine-tuning, suggesting a lack of logical reasoning and overtraining if contextual documents are not provided. The same results can be seen in the
BLEU and ROUGE scores, where using the original dataset for training show the best results when the context is provided, with the best BLEU scores of 0.489 and 0.463, and ROUGE scores of 0.494 and 0.485 for Llama and Mistral, respectively. Conversely, using the generated dataset for training shows the best results when no context is provided, with the best BLEU scores of 0.116 and 0.172, and ROUGE scores of 0.217 and 0.260 for Llama and Mistral respectively. The BLEU and ROUGE scores are significantly lower when no context is provided, indicating a lack of semantic and structural similarities to the original labelled answers, though the BERT F1 score remains similarly high.

\section{Discussion}

\subsection{Limitations}\label{sec:limitations}
The automated QA generation and subsequent fine-tuning of the generated dataset has only been tested on the TechQA dataset. While this dataset closely aligns with the use of technical documentation in QA tasks, it is not known how well the QA dataset generation and fine-tuning apply to other forms of data and documentation. In the same manner the in-context results rely on a user already familiar with the material to a level they can find the document with the correct QA in it or that there is a retrieval system in place capable of performing the same task perfectly. 

Two smaller LLMs, Llama-3-8b and Mistral-7b-v0.3, are fine-tuned and tested, which demonstrate the effectiveness of fine-tuning on a generated dataset in cases with limited compute resources. The use of QLoRA further reduces the computational requirements of the fine-tuning process. These choices in smaller models and quantisation reduces the computation required but also reduces the quality of answers, leading to lower evaluation metrics particularly for the BLEU and ROUGE scores. Larger models with $>8B$ parameters, commercial closed sourced models, and un-quantised full fine-tuning have not been tested, which may demonstrate higher quality answers without the need for fine-tuning on either the original or generated dataset.

\subsection{Societal Impact}\label{sec:impact}
Our work can enhance the creation of QA systems through the reduced human labour in creating QA datasets for LLM fine-tuning. Annotation time and costs have been estimated in the region of 3 person-weeks for 20,000 annotations, which in a technical context requires highly trained annotators \cite{hosseini2024retrieveannotateevaluaterepeat}. In particular, LLM users who wish to develop their own models based on private documentation would not need to manually create datasets. The use of synthetic datasets contributes to further AI development, rapidly increasing the amount of data available for training and testing.

There are a few risks to creating and using synthetic datasets. First is the risk of job displacements within technical and professional fields, replacing domain-specific expertise with AI expertise. Domain expertise is necessary in the creation of such QA systems, for evaluation, management, and further improvements, but there is a risk that experts are seen as redundant. Secondly, the overuse and reliance on synthetic data could lead to bias during the generation stage, or reduced LLM performance. Further studies on the balance of datasets curated by humans and synthetically generated datasets should be conducted to understand the limitations of using synthetic data. Moreover, the human cost of creating QA datasets is shifted to computational costs of generating the data and training, as synthetic datasets can be much larger than human-labelled datasets.

\section{Conclusion}
QA systems are increasingly used in a range of applications. The generation of the data to train these models, however, can be expensive, especially if the documentation is technical. In this paper, we introduce a framework to generate synthetic QA datasets with LLMs and show that fine-tuning using these synthetic datasets improves performance. However, we show that training on the original dataset remains the best approach to find answers to questions when the correct contextual documents are provided. Notably, we demonstrate that LLMs trained on the generated datasets are able to better learn on the technical documentation, providing higher BERT F1, BLEU, and ROUGE scores of 0.843, 0.116, and 0.217 respectively when trained on the generated dataset compared to scores of 0.831, 0.072, and 0.194 when trained on the original dataset for Llama-3-8b, and BERT F1, BLEU, and ROUGE scores 0.858, 0.172, and 0.260 compared to scores of 0.836, 0.083, and 0.139 for Mistral-7b-v0.3.


{\small
\bibliographystyle{plainnat}
\bibliography{main}
}

\appendix

\section{Table of results}
The full tables of results are displayed here, with every model, prompting strategy and quantitative metric.
The best result for each metric and model is highlighted in bold.

\begin{table}[!ht]
  \centering
  \caption{With Context Results on TechQA.}
  \label{tab:techqa}
  \resizebox{\linewidth}{!}{\begin{tabular}{c||c|c|c|c|c|c|c|c|c} \toprule
    Model & Training Set & Training Shot & Inference Shot & EM & Precision & Recall & F1 & BLEU & ROUGE \\ \hline
    \multirow{21}{*}{Llama-3-8b} & \multirow{3}{*}{no training} & \multirow{3}{*}{-} & 0-shot & 0.000 & 0.810 & 0.864 & 0.835 & 0.069 & 0.208 \\
    & & & 1-shot & 0.000 & 0.879 & 0.885 & 0.881 & 0.153 & 0.270 \\
    & & & 5-shot & 0.000 & 0.879 & 0.887 & 0.883 & 0.276 & 0.449 \\ \cline{2-10} 
    & \multirow{9}{*}{original} & \multirow{3}{*}{0-shot} & 0-shot & 0.000 & 0.866 & 0.893 & 0.878 & 0.087 & 0.377 \\
    & & & 1-shot & 0.154 & 0.881 & 0.888 & 0.883 & 0.140 & 0.425 \\
    & & & 5-shot & \textbf{0.220} & \textbf{0.898} & 0.884 & 0.890 & 0.141 & 0.431 \\ \cline{3-10}
    & & \multirow{3}{*}{1-shot} & 0-shot & 0.165 & 0.852 & 0.875 & 0.862 & 0.074 & 0.331 \\
    & & & 1-shot & 0.000 & 0.874 & 0.891 & 0.881 & 0.135 & 0.418 \\
    & & & 5-shot & 0.176 & \textbf{0.898} & 0.895 & 0.895 & 0.158 & 0.484 \\ \cline{3-10}
    & & \multirow{3}{*}{5-shot} & 0-shot & 0.066 & 0.789 & 0.888 & 0.835 & 0.062 & 0.177 \\
    & & & 1-shot & 0.000 & 0.873 & 0.890 & 0.880 & 0.354 & 0.423 \\
    & & & 5-shot & 0.077 & 0.894 & \textbf{0.907} & \textbf{0.899} & \textbf{0.489} & \textbf{0.494} \\ \cline{2-10}
    & \multirow{9}{*}{generated} & \multirow{3}{*}{0-shot} & 0-shot & 0.0 & 0.834 & 0.868 & 0.850 & 0.092 & 0.262 \\
    & & & 1-shot & 0.110 & 0.838 & 0.884 & 0.859 & 0.148 & 0.272 \\
    & & & 5-shot & 0.000 & 0.835 & 0.888 & 0.859 & 0.117 & 0.267 \\ \cline{3-10}
    & & \multirow{3}{*}{1-shot} & 0-shot & 0.000 & 0.832 & 0.866 & 0.848 & 0.181 & 0.285 \\
    & & & 1-shot & 0.022 & 0.880 & 0.899 & 0.889 & 0.252 & 0.429 \\
    & & & 5-shot & 0.000 & 0.831 & 0.871 & 0.850 & 0.088 & 0.266 \\ \cline{3-10}
    & & \multirow{3}{*}{5-shot} & 0-shot & 0.000 & 0.848 & 0.870 & 0.858 & 0.159 & 0.306 \\
    & & & 1-shot & 0.011 & 0.843 & 0.880 & 0.859 & 0.176 & 0.276 \\
    & & & 5-shot & 0.000 & 0.835 & 0.872 & 0.852 & 0.095 & 0.263 \\ \hline 
    \multirow{21}{*}{Mistral-7b-v0.3} & \multirow{3}{*}{no training} & \multirow{3}{*}{-} & 0-shot & 0.000 & 0.818 & 0.862 & 0.839 & 0.071 & 0.232 \\ 
    & & & 1-shot & 0.000 & 0.824 & 0.868 & 0.844 & 0.081 & 0.259 \\
    & & & 5-shot & 0.000 & 0.835 & 0.877 & 0.854 & 0.122 & 0.269 \\ \cline{2-10} 
    & \multirow{9}{*}{original} & \multirow{3}{*}{0-shot} & 0-shot & 0.000 & 0.899 & 0.898 & 0.897 & 0.416 & 0.479 \\ 
    & & & 1-shot & 0.000 & 0.890 & 0.900 & 0.894 & 0.417 & 0.475 \\
    & & & 5-shot & 0.000 & 0.885 & 0.898 & 0.891 & 0.393 & 0.448 \\ \cline{3-10}
    & & \multirow{3}{*}{1-shot} & 0-shot & 0.000 & 0.888 & 0.891 & 0.888 & 0.377 & 0.439 \\
    & & & 1-shot & 0.000 & 0.901 & 0.894 & 0.897 & 0.376 & 0.468 \\
    & & & 5-shot & 0.000 & 0.903 & 0.903 & \textbf{0.902} & 0.460 & \textbf{0.485} \\ \cline{3-10}
    & & \multirow{3}{*}{5-shot} & 0-shot & 0.000 & 0.885 & 0.890 & 0.887 & 0.373 & 0.430 \\
    & & & 1-shot & 0.000 & 0.903 & 0.896 & 0.899 & 0.388 & 0.476 \\
    & & & 5-shot & 0.000 & \textbf{0.904} & 0.903 & \textbf{0.902} & \textbf{0.463} & 0.484 \\ \cline{2-10}
    & \multirow{9}{*}{generated} & \multirow{3}{*}{0-shot} & 0 shot & 0.000 & 0.818 & 0.869 & 0.841 & 0.090 & 0.258 \\
    & & & 1-shot & 0.000 & 0.828 & 0.892 & 0.858 & 0.137 & 0.292 \\
    & & & 5-shot & 0.000 & 0.822 & 0.871 & 0.845 & 0.095 & 0.250 \\ \cline{3-10}
    & & \multirow{3}{*}{1-shot} & 0 shot & 0.000 & 0.819 & 0.869 & 0.842 & 0.084 & 0.232 \\
    & & & 1 shot & \textbf{0.011} & 0.854 & 0.887 & 0.869 & 0.175 & 0.365 \\
    & & & 5 shot & 0.000 & 0.853 & 0.883 & 0.867 & 0.021 & 0.075 \\ \cline{3-10}
    & & \multirow{3}{*}{5-shot} & 0 shot & 0.000 & 0.861 & 0.902 & 0.880 & 0.235 & 0.412 \\
    & & & 1 shot & 0.000 & 0.861 & \textbf{0.907} & 0.882 & 0.232 & 0.422 \\ 
    & & & 5 shot & 0.000 & 0.828 & 0.874 & 0.850 & 0.113 & 0.275 \\ \bottomrule
  \end{tabular}}
\end{table}

\begin{table}[!ht]
  \centering
  \caption{No Context Results on TechQA.}
  \label{tab:techqa-no-context}
  \resizebox{\linewidth}{!}{\begin{tabular}{c||c|c|c|c|c|c|c|c|c} \toprule
    Model & Training Set & Training Shot & Inference Shot & EM & Precision & Recall & F1 & BLEU & ROUGE \\ \hline
    \multirow{21}{*}{Llama-3-8b} & \multirow{3}{*}{no training} & \multirow{3}{*}{-} & 0-shot & 0.000 & 0.813 & 0.811 & 0.811 & 0.013 & 0.069 \\
    & & & 1-shot & 0.000 & 0.826 & 0.837 & 0.831 & 0.021 & 0.141 \\
    & & & 5-shot & 0.000 & 0.830 & 0.840 & 0.835 & 0.034 & 0.162 \\ \cline{2-10} 
    & \multirow{9}{*}{original} & \multirow{3}{*}{0-shot} & 0-shot & 0.000 & 0.808 & 0.837 & 0.822 & 0.032 & 0.128 \\ 
    & & & 1-shot & 0.154 & 0.810 & 0.834 & 0.821 & 0.028 & 0.127 \\
    & & & 5-shot & \textbf{0.220} & 0.821 & 0.843 & 0.831 & 0.072 & 0.165 \\ \cline{3-10}
    & & \multirow{3}{*}{1-shot} & 0-shot & 0.165 & 0.800 & 0.832 & 0.815 & 0.014 & 0.153 \\
    & & & 1-shot & 0.000 & 0.806 & 0.832 & 0.818 & 0.025 & 0.152 \\
    & & & 5-shot & 0.176 & 0.814 & 0.837 & 0.825 & 0.036 & 0.164 \\ \cline{3-10}
    & & \multirow{3}{*}{5-shot} & 0-shot & 0.066 & 0.802 & 0.834 & 0.817 & 0.028 & 0.146 \\
    & & & 1-shot & 0.000 & 0.814 & 0.832 & 0.823 & 0.020 & 0.158 \\
    & & & 5-shot & 0.077 & 0.818 & 0.840 & 0.828 & 0.057 & 0.180 \\ \cline{2-10}
    & \multirow{9}{*}{generated} & \multirow{3}{*}{0-shot} & 0-shot & 0.000 & 0.802 & 0.833 & 0.816 & 0.022 & 0.121 \\
    & & & 1-shot & 0.110 & 0.805 & 0.833 & 0.818 & 0.013 & 0.116 \\
    & & & 5-shot & 0.000 & \textbf{0.836} & 0.851 & \textbf{0.843} & \textbf{0.116} & 0.194 \\ \cline{3-10}
    & & \multirow{3}{*}{1-shot} & 0-shot & 0.000 & 0.802 & 0.835 & 0.817 & 0.058 & 0.128 \\
    & & & 1-shot & 0.022 & 0.834 & \textbf{0.853} & \textbf{0.843} & 0.105 & \textbf{0.217} \\
    & & & 5-shot & 0.000 & 0.830 & 0.836 & 0.833 & 0.041 & 0.151 \\ \cline{3-10}
    & & \multirow{3}{*}{5-shot} & 0-shot & 0.000 & 0.756 & 0.828 & 0.790 & 0.026 & 0.086 \\
    & & & 1-shot & 0.011 & 0.822 & 0.834 & 0.827 & 0.034 & 0.133 \\
    & & & 5-shot & 0.000 & 0.834 & 0.834 & 0.834 & 0.030 & 0.139 \\ \hline 
    \multirow{21}{*}{Mistral-7b-v0.3} & \multirow{3}{*}{no training} & \multirow{3}{*}{-} & 0-shot & 0.000 & 0.802 & 0.835 & 0.818 & 0.016 & 0.126 \\ 
    & & & 1-shot & 0.000 & 0.810 & 0.839 & 0.824 & 0.021 & 0.144 \\
    & & & 5-shot & 0.000 & 0.807 & 0.838 & 0.822 & 0.028 & 0.145 \\ \cline{2-10} 
    & \multirow{9}{*}{original} & \multirow{3}{*}{0-shot} & 0-shot & 0.000 & 0.820 & 0.838 & 0.828 & 0.036 & 0.133 \\ 
    & & & 1-shot & 0.000 & 0.831 & 0.833 & 0.831 & 0.047 & 0.138 \\
    & & & 5-shot & 0.000 & 0.839 & 0.835 & 0.836 & 0.062 & 0.107 \\ \cline{3-10}
    & & \multirow{3}{*}{1-shot} & 0-shot & 0.000 & 0.802 & 0.834 & 0.817 & 0.020 & 0.133 \\
    & & & 1-shot & 0.000 & 0.814 & 0.832 & 0.822 & 0.030 & 0.107 \\
    & & & 5-shot & 0.000 & 0.830 & 0.835 & 0.832 & 0.083 & 0.139 \\ \cline{3-10}
    & & \multirow{3}{*}{5-shot} & 0-shot & 0.000 & 0.807 & 0.835 & 0.820 & 0.022 & 0.112 \\
    & & & 1-shot & 0.000 & 0.816 & 0.832 & 0.822 & 0.029 & 0.131 \\
    & & & 5-shot & 0.000 & 0.826 & 0.835 & 0.830 & 0.073 & 0.134 \\ \cline{2-10}
    & \multirow{9}{*}{generated} & \multirow{3}{*}{0-shot} & 0 shot & 0.000 & 0.806 & 0.835 & 0.820 & 0.014 & 0.128 \\
    & & & 1-shot & 0.000 & 0.843 & 0.855 & 0.849 & 0.139 & 0.224 \\
    & & & 5-shot & 0.000 & 0.850 & 0.856 & 0.853 & 0.164 & 0.244 \\ \cline{3-10}
    & & \multirow{3}{*}{1-shot} & 0 shot & 0.000 & 0.788 & 0.828 & 0.807 & 0.011 & 0.103 \\
    & & & 1 shot & \textbf{0.011} & \textbf{0.855} & \textbf{0.861} & \textbf{0.858} & 0.170 & 0.170 \\
    & & & 5 shot & 0.000 & 0.853 & 0.858 & 0.855 & \textbf{0.172} & \textbf{0.260} \\ \cline{3-10}
    & & \multirow{3}{*}{5-shot} & 0 shot & 0.000 & 0.760 & 0.819 & 0.787 & 0.014 & 0.094 \\
    & & & 1 shot & 0.000 & 0.848 & 0.855 & 0.851 & 0.165 & 0.242 \\ 
    & & & 5 shot & 0.000 & 0.849 & 0.852 & 0.850 & 0.150 & 0.241 \\ \bottomrule
  \end{tabular}}
\end{table}

\FloatBarrier

\section*{NeurIPS Paper Checklist}

\begin{enumerate}

\item {\bf Claims}
    \item[] Question: Do the main claims made in the abstract and introduction accurately reflect the paper's contributions and scope?
    \item[] Answer: Yes
    \item[] Justification: The claims made in the abstract and introduction are demonstrated in the results section.

\item {\bf Limitations}
    \item[] Question: Does the paper discuss the limitations of the work performed by the authors?
    \item[] Answer: Yes
    \item[] Justification: The limitations of our methods are discussed in section \ref{sec:limitations} of the discussion.

\item {\bf Theory assumptions and proofs}
    \item[] Question: For each theoretical result, does the paper provide the full set of assumptions and a complete (and correct) proof?
    \item[] Answer: NA
    \item[] Justification: The paper does not propose theoretical assumptions and does not include theoretical results

\item {\bf Experimental result reproducibility}
    \item[] Question: Does the paper fully disclose all the information needed to reproduce the main experimental results of the paper to the extent that it affects the main claims and/or conclusions of the paper (regardless of whether the code and data are provided or not)?
    \item[] Answer: Yes
    \item[] Justification: The process for QA generation and fine-tuning details are described in the methodology. The hyperparameters, models, and frameworks used for fine-tuning are also described. Furthermore, a zipfile with the code is provided as supplementary material, and a link to the repository can be provided after submission to maintain anonymity.
    \item[] Guidelines:
    \begin{itemize}
        \item The answer NA means that the paper does not include experiments.
        \item If the paper includes experiments, a No answer to this question will not be perceived well by the reviewers: Making the paper reproducible is important, regardless of whether the code and data are provided or not.
        \item If the contribution is a dataset and/or model, the authors should describe the steps taken to make their results reproducible or verifiable. 
        \item Depending on the contribution, reproducibility can be accomplished in various ways. For example, if the contribution is a novel architecture, describing the architecture fully might suffice, or if the contribution is a specific model and empirical evaluation, it may be necessary to either make it possible for others to replicate the model with the same dataset, or provide access to the model. In general, releasing code and data is often one good way to accomplish this, but reproducibility can also be provided via detailed instructions for how to replicate the results, access to a hosted model (e.g., in the case of a large language model), releasing of a model checkpoint, or other means that are appropriate to the research performed.
        \item While NeurIPS does not require releasing code, the conference does require all submissions to provide some reasonable avenue for reproducibility, which may depend on the nature of the contribution. For example
        \begin{enumerate}
            \item If the contribution is primarily a new algorithm, the paper should make it clear how to reproduce that algorithm.
            \item If the contribution is primarily a new model architecture, the paper should describe the architecture clearly and fully.
            \item If the contribution is a new model (e.g., a large language model), then there should either be a way to access this model for reproducing the results or a way to reproduce the model (e.g., with an open-source dataset or instructions for how to construct the dataset).
            \item We recognize that reproducibility may be tricky in some cases, in which case authors are welcome to describe the particular way they provide for reproducibility. In the case of closed-source models, it may be that access to the model is limited in some way (e.g., to registered users), but it should be possible for other researchers to have some path to reproducing or verifying the results.
        \end{enumerate}
    \end{itemize}

\item {\bf Open access to data and code}
    \item[] Question: Does the paper provide open access to the data and code, with sufficient instructions to faithfully reproduce the main experimental results, as described in supplemental material?
    \item[] Answer: Yes
    \item[] Justification: The data and code are provided as supplementary material.

\item {\bf Experimental setting/details}
    \item[] Question: Does the paper specify all the training and test details (e.g., data splits, hyperparameters, how they were chosen, type of optimizer, etc.) necessary to understand the results?
    \item[] Answer: Yes
    \item[] Justification: Hyperparameters for fine-tuning, the data splits, how they were chosen, and the type of optimiser are detailed in the methodology and results sections, with a URL to the code also provided.

\item {\bf Experiment statistical significance}
    \item[] Question: Does the paper report error bars suitably and correctly defined or other appropriate information about the statistical significance of the experiments?
    \item[] Answer: No
    \item[] Justification: Variables which can affect our results is only the fine-tuning and inference random seed. Due to limited computational resources, we omitted error bars from repeated runs with different random seeds. Nevertheless, we find that all inference runs are stable, providing consistent results across runs with the same hyperparameters.

\item {\bf Experiments compute resources}
    \item[] Question: For each experiment, does the paper provide sufficient information on the computer resources (type of compute workers, memory, time of execution) needed to reproduce the experiments?
    \item[] Answer: Yes
    \item[] Justification: The compute resources required is detailed in the results section.
    
\item {\bf Code of ethics}
    \item[] Question: Does the research conducted in the paper conform, in every respect, with the NeurIPS Code of Ethics \url{https://neurips.cc/public/EthicsGuidelines}?
    \item[] Answer: Yes
    \item[] Justification: The research we conducted in the paper conform, in every respect, with the NeurIPS Code of Ethics. 

\item {\bf Broader impacts}
    \item[] Question: Does the paper discuss both potential positive societal impacts and negative societal impacts of the work performed?
    \item[] Answer: Yes
    \item[] Justification: Positive and negative societal impacts are discussed in section \ref{sec:impact}
    
\item {\bf Safeguards}
    \item[] Question: Does the paper describe safeguards that have been put in place for responsible release of data or models that have a high risk for misuse (e.g., pretrained language models, image generators, or scraped datasets)?
    \item[] Answer: NA
    \item[] Justification: The paper poses no such risks.

\item {\bf Licenses for existing assets}
    \item[] Question: Are the creators or original owners of assets (e.g., code, data, models), used in the paper, properly credited and are the license and terms of use explicitly mentioned and properly respected?
    \item[] Answer: Yes
    \item[] Justification: The licenses of the datasets and models used are included in the text where they are introduced.

\item {\bf New assets}
    \item[] Question: Are new assets introduced in the paper well documented and is the documentation provided alongside the assets?
    \item[] Answer: NA
    \item[] Justification: We do not release new assets in the paper.

\item {\bf Crowdsourcing and research with human subjects}
    \item[] Question: For crowdsourcing experiments and research with human subjects, does the paper include the full text of instructions given to participants and screenshots, if applicable, as well as details about compensation (if any)? 
    \item[] Answer: NA
    \item[] Justification: The paper does not involve crowdsourcing nor research with human subjects

\item {\bf Institutional review board (IRB) approvals or equivalent for research with human subjects}
    \item[] Question: Does the paper describe potential risks incurred by study participants, whether such risks were disclosed to the subjects, and whether Institutional Review Board (IRB) approvals (or an equivalent approval/review based on the requirements of your country or institution) were obtained?
    \item[] Answer: NA
    \item[] Justification: The paper does not involve crowdsourcing nor research with human subjects

\item {\bf Declaration of LLM usage}
    \item[] Question: Does the paper describe the usage of LLMs if it is an important, original, or non-standard component of the core methods in this research? Note that if the LLM is used only for writing, editing, or formatting purposes and does not impact the core methodology, scientific rigorousness, or originality of the research, declaration is not required.
    \item[] Answer: Yes
    \item[] Justification: The usage of LLMs to generate a synthetic QA dataset is described fully in the methodology section

\end{enumerate}

\end{document}